# A New Calibration Method for Industrial Robot Based on Step-Size Levenberg-Marquardt Algorithm

Zhibin Li, Shuai Li, *Senior Member*, *IEEE*, and Xin Luo, *Senior Member*, *IEEE*

*Abstract*—Industrial robots play a vital role in automatic production, which have been widely utilized in industrial production activities, like handling and welding. However, due to an uncalibrated robot with machining tolerance and assembly tolerance, it suffers from low absolute positioning accuracy, which cannot satisfy the requirements of high-precision manufacture. To address this hot issue, we propose a novel calibration method based on an unscented Kalman filter and variable step-size Levenberg-Marquardt algorithm. This work has three ideas: a) proposing a novel variable step-size Levenberg-Marquardt algorithm to addresses the issue of local optimum in a Levenberg-Marquardt algorithm; b) employing an unscented Kalman filter to reduce the influence of the measurement noises; and c) developing a novel calibration method incorporating an unscented Kalman filter with a variable step-size Levenberg-Marquardt algorithm. Furthermore, we conduct enough experiments on an ABB IRB 120 industrial robot. From the experimental results, the proposed method achieves much higher calibration accuracy than some state-of-the-art calibration methods. Hence, this work is an important milestone in the field of robot calibration.

## I. INTRODUCTION

INDUSTRIAL robots are the major equipment of advanced manufacture, whose widely industrial applications are the critical symbol to weigh the development level of the technological innovation and high-precision manufacture. Vigorously developing the robotic industry has an excellent effect on accelerating advanced manufacture [1-5], which promotes the industrial upgrading and improves our daily life.

With the extensively expansion of the application of industrial robots, which demands to satisfy the complex tasks with higher precision requirements [5-10]. Generally, an industrial robots enjoys its greatly high repetitive positioning accuracy, while the absolute positioning error of an uncalibrated robot might reach several millimeters, which cannot satisfy the demands of high-precision off-line programming and intelligent application. Hence, to improve the performance of the robot, it is necessary to calibrate the kinematic parameters [11-15].

Aiming at addressing the issue of robot calibration, researchers have made numerous explorations [16-20]. For instance, Nguyen *et al*. [2] incorporated an extended Kalman filter (EKF) algorithm and an artificial neural network to calibrate the serial PUMA and HH800 robots. The EKF algorithm was adopted to identify the geometric parameters, then the artificial neural network was utilized to compensate the non-geometric errors, thereby greatly enhancing positioning accuracy of the robots. Wang *et al*. [4] proposed a calibration method via integrating back propagation (BP) network and analytical method, they adopted limited parameters calibration, which achieved excellent experimental results after calibration. Wang *et al*. [6] developed a multilayer perceptron neural network (MLPNN) combining with beetle swarm optimization algorithm. Extensive experiments implemented on a SIASUN SR210D robot manipulator demonstrated that this method had a faster convergence rate and higher calibration accuracy. Gan *et al*. [1] presented a robot calibration method with four drawstring displacement sensors, which adopted Levenberg-Marquardt (LM) algorithm to accurately identify the kinematic parameter errors for an EFORT ER3A robot. Actually, the mentioned calibration methods improve the robot calibration accuracy. However, they frequently encounter falling into local optimum, which exceedingly harms the calibration accuracy.

To address this thorny issue, we propose a new variable step-size Levenberg-Marquardt algorithm, which is motivated by a beetle antennae search (BAS) algorithm. Additionally, it improves the ability of the LM algorithm to jump out of the local optimum [21-25]. Moreover, this algorithm has the advantages of easy implementation, high efficiency, strong robustness, which is appropriate for application in occasions with high accuracy requirements of the robot [26-30].

Due to the measurement noises in robot calibration process, some calibration methods cannot accurately identify the positioning errors, which significantly harm its calibration accuracy. For addressing this thorny issue, this paper proposes a novel calibration method based on an unscented Kalman filter (UKF) algorithm and a variable step-size Levenberg-Marquardt (SLM) algorithm. The main contributions of this work are summarized below:
a) A new SLM algorithm with variable step-size term, which improve the search ability of a LM algorithm;

b) An unscented Kalman filter is utilized to suppress the measurement noises;
c) A novel calibration method based on an UKF algorithm and a SLM algorithm, which obtains much higher calibration accuracy than it peers do.

Section II presents the preliminaries. Section III describes the UKF and SLM algorithm for robot calibration. The experimental comparisons and verifications are given in Section IV. Finally, Section V concludes this paper with some conclusions.

## II. Preliminaries

### A. Kinematic and Error Model for Calibration

Robot calibration mainly includes four parts: modeling, measurement, parameter identification and error compensation [51-55]. The modeling has a huge influence on the robot positioning error. Recently, the classical kinematic model of a serial robot is DH model. Hence, we adopt DH model to establish the kinematic model [35-40].

According to the definition of DH model [41-46], the link transformation matrix of its adjacent joints can be expressed as

$$K_i = \begin{bmatrix} \cos\theta_i & -\sin\theta_i \cos\alpha_i & \sin\theta_i \sin\alpha_i & a_i \cos\theta_i \\ \sin\theta_i & \cos\theta_i \cos\alpha_i & -\cos\theta_i \sin\alpha_i & a_i \sin\theta_i \\ 0 & \sin\alpha_i & \cos\alpha_i & d_i \\ 0 & 0 & 0 & 1 \end{bmatrix}, \quad (1)$$

where $a_i$, $d_i$, $\theta_i$ and $\alpha_i$ are link parameters of the robot, which contains the link length, link offset, joint angle, link twist angle [61-65]. $K$ is link transformation matrix. Then the transformation matrix from the robotic base coordinate system to the end coordinate system is given as

$$K_6^0 = K_1^0 K_2^1 K_3^2 K_4^3 K_5^4 K_6^5. \quad (2)$$

Considering the kinematic parameter errors of each link, then the deviation of robotic transformation matrix can be represented as

$$dK = K_R - K_6^0, \quad (3)$$

where $dK$ represents the pose deviation of the robot. Based on equation (3), $dK$ can be written as:

$$dK_i = \frac{\partial K_i}{\partial \alpha_i} d\alpha_i + \frac{\partial K_i}{\partial a_i} da_i + \frac{\partial K_i}{\partial d_i} dd_i + \frac{\partial K_i}{\partial \theta_i} d\theta_i. \quad (4)$$

Combining (3) and (4), the pose error model of robot end-effector is as follow:

$$e = \begin{bmatrix} J_1 & J_2 & J_3 & J_4 \end{bmatrix} \begin{bmatrix} \Delta a \\ \Delta d \\ \Delta \alpha \\ \Delta \theta \end{bmatrix} = JX, \quad (5)$$

$J$ represents the differential identification Jacobian matrix, $X$ is the deviations of the robot kinematic parameters.

Based on the principle of robot calibration [31-35], the difference between the measurement cable length and the nominal cable length is approximately equal to the robot positioning error, whose objective function is given as

$$f(X) = \min\left[\frac{1}{m}\sum_{i=1}^{m}(Z_i - Z'_i)^2\right], \quad (6)$$

where $Z'_i$ and $Z_i$ are the nominal cable length, measured cable length, respectively. $m$ is the number of sampling points.

Commonly, the nominal cable length can be calculated by

$$Z'_i = \sqrt{(P_i - P_0)^2}, \quad (7)$$

$P_i$ is the nominal position of robot end-effector, $P_0$ is the position of fixed point.

## III. Robot Calibration Model

### A. SLM Algorithm

#### 1) Least-squares Algorithm

Least-squares algorithm is the most commonly employed to address the issue of curve-fitting, which has been widely applied in finance, electronics, medical care and other fields. For the robot error identification model (5), the least-squares solution of robot calibration can be calculated as the following formula when the column vector of the identification Jacobian matrix $J$ is full rank.

$$X = (J^T J)^{-1} \cdot J^T \cdot E, \quad (8)$$

where $E = Z_i - Z'_i$, $X$ is the deviations of kinematic parameters.

*2) LM Algorithm*

In general, the robot error model has redundant parameters, which frequently encounters the singular phenomena. In other words, some parameters cannot be identified [45-49]. The ordinary least-squares algorithm may cause solution errors, thus LM algorithm is adopted to address this issue, whose solution is given as

$$X = \left(J^T J + \lambda I\right)^{-1} \cdot J^T \cdot E, \qquad (9)$$

$\lambda$ represents the learning rate, $I$ is the unit matrix.

*3) SLM Algorithm*

To improve the search ability of LM algorithm, inspired by a BAS algorithm, we incorporate variable step-size term into its evaluation rule. This method enormously improves its ability to jump out of local optimum in high-dimensional space. Meanwhile, it also has a higher calibration accuracy and calculation efficiency, whose updating rule can be obtained as

$$\begin{cases} X = \left(J^T J + \lambda I\right)^{-1} \cdot J^T \cdot E \cdot \delta_t, \\ \delta_{t+1} = \delta_t \cdot \mu, \end{cases} \qquad (10)$$

where $\delta$ is the step size of searching, $\mu \in (0,1)$.

*B. UKF Algorithm*

Considering the principle of UKF algorithm, we adopt an unscented transformation to the sigma vector. The posteriori sigma vector $X'_{i,k|k-1}$ and the posteriori measurement state $Y'_{i,k|k-1}$ are given as

$$\begin{cases} X'_{i,k|k-1} = \Phi_k\left(X'_{i,k-1}\right), \\ Y'_{i,k|k-1} = H_k\left(X'_{i,k|k-1}\right). \end{cases} \qquad (11)$$

Based on (11), we can achieve the covariance $P_k^-$ and priori state $x_k^-$, $W_i$ is the weight of sigma point $i$.

$$\begin{cases} x_k^- = \sum_{i=0}^{N+1} W_i X'_{i,k|k-1}, \\ P_k^- = \sum_{i=0}^{N+1} W_i \left(X'_{i,k|k-1} - x_k^-\right)\left(X'_{i,k|k-1} - x_k^-\right)^T. \end{cases} \qquad (12)$$

Then the mean $\hat{y}$ and covariance $S$ of the observation model are written as

$$\begin{cases} \hat{y}_{k|k-1} = \sum_{i=0}^{N+1} W_i Y'_{i,k|k-1}, \\ S_{k|k-1} = \sum_{i=0}^{N+1} W_i \left(Y'_{i,k|k-1} - \hat{y}_{k|k-1}\right)\left(Y'_{i,k|k-1} - \hat{y}_{k|k-1}\right)^T. \end{cases} \qquad (13)$$

Additionally, the Kalman gain $\rho$ and the cross covariance $P_{xy}$ can be represented as

$$\begin{cases} P_{xy,k|k-1} = \sum_{i=0}^{N+1} W_i \left(X'_{i,k|k-1} - x_k^-\right)\left(Y'_{i,k|k-1} - \hat{y}_{k|k-1}\right)^T, \\ \rho = P_{xy,k|k-1} S_{k|k-1}^{-1}. \end{cases} \qquad (14)$$

Lastly, the estimation value of $X$ and the updating covariance $P_k$ is calculated by

$$\begin{cases} x_k = x_{k|k-1}^- + \rho\left(Y_k - \hat{y}_{k|k-1}\right), \\ P_k = P_{k|k-1}^- - \rho S_{k|k-1} \rho^T, \end{cases} \qquad (15)$$

$Y_k$ represents the current set of observations.

## IV. Experimental Results

*A. General Settings*

*1) Evaluation Metrics:* We adopt the root mean squared error (RMSE), average error (Std) and the maximum error (Max) as the evaluation metrics to validate the performance of our proposed method [18], whose calculation formula can be expressed as

$$Max = \max\left\{\sqrt{\left(Z_i - Z'_i\right)^2}\right\}, Std = \frac{1}{m}\sum_{i=1}^{m}\sqrt{\left(Z_i - Z'_i\right)^2},\ RMSE = \sqrt{\frac{1}{m}\sum_{i=1}^{m}\left(Z_i - Z'_i\right)^2},\ i = 1,2,\cdots m. \qquad (16)$$

*2) Dataset:* To obtain 120 diversified samples, we should collect the samples in different positions. Thereafter, this dataset is transferred by a RS485 communication module. Lastly, we employ the proposed method to identify robot kinematic parameter errors. This dataset is publicly available on https://github.com/Lizhibing1490183152/RobotCali.

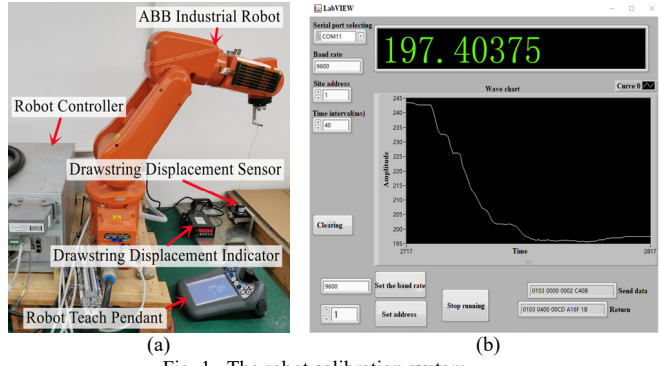

Fig. 1. The robot calibration system.

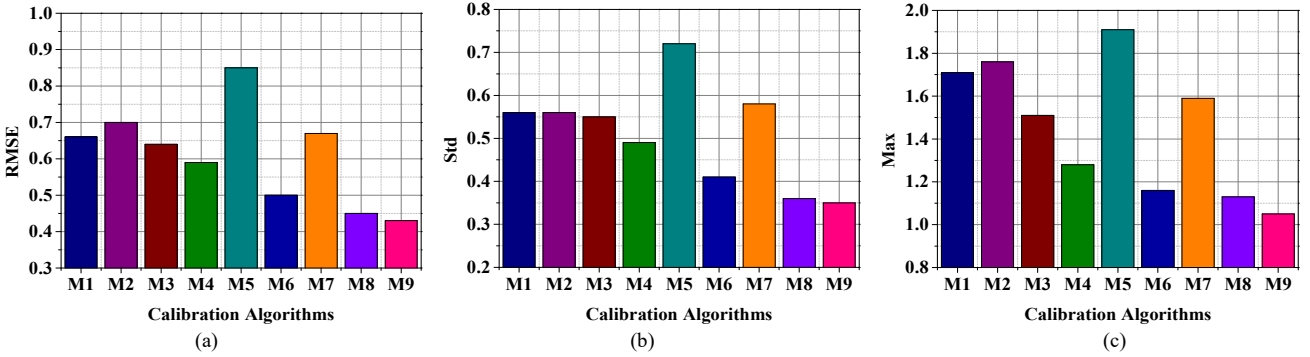

Fig. 2. Calibration accuracy of compared methods.

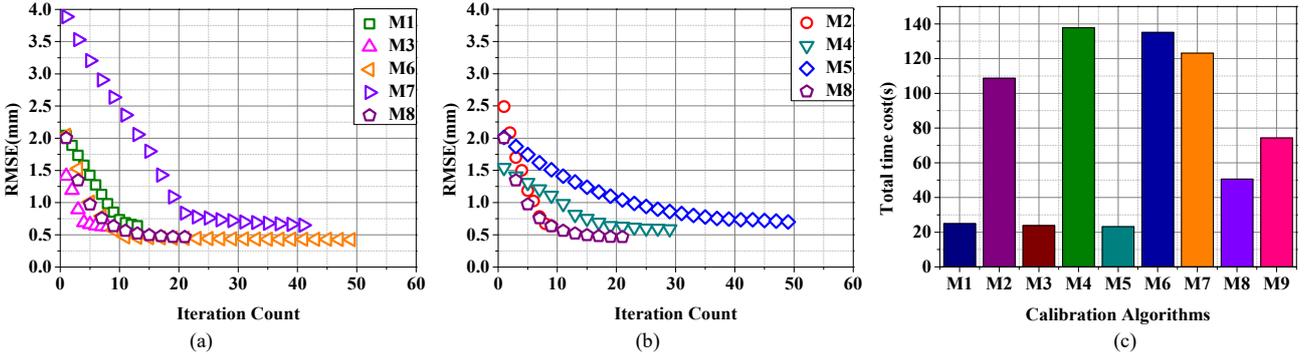

Fig. 3. Training curves and total time cost of calibration methods.

TABLE I
COMPARED CALIBRATION METHODS.

| Method | Description |
|---|---|
| M1 | The extended Kalman filter (EKF) algorithm, which is presented in [8]. It can address the non-Gaussian noise in calibration system. |
| M2 | The Beetle antennae search (BAS) algorithm [6], which simulates the food searching behavior of beetles. |
| M3 | The unscented Kalman filter algorithm (UKF), which is developed in [12]. |
| M4 | The particle swarm optimization algorithm (PSO) [28], which simulates the foraging behavior of birds to achieve optimization solution. |
| M5 | The Radial basis function neural network (RBF) proposed in [13], which is used to identify the non-geometric errors of the robot. |
| M6 | The Levenberg-Marquardt (LM) algorithm proposed in [1] that is adopted to identify the deviation of kinematic parameters. |
| M7 | The differential evolution algorithm (DE) [18]. It is a multi-objective optimization algorithm, which can accurately calibrate the robot. |
| M8 | The variable step-size Levenberg-Marquardt (SLM) algorithm, which improves the search ability of LM algorithm. |
| M9 | The proposed UKF-SLM algorithm, which can significantly improve the calibration performance of the robot. |

*3) Experimental Platform:* We design an experimental platform as shown in Fig. 1(a), which includes an ABB IRB120 industrial robot, a drawstring displacement sensor and a drawstring displacement indicator.

*4) Experimental Process:* In this work, 120 measurement points are selected in the workspace of an ABB IRB120 industrial robot. Furthermore, a data acquisition program is developed based on the LabVIEW software, which is shown in Fig. 1(b).

B. *Comparison With State-of-the-Art Methods*

In the experiment, we compare the performance of the proposed method against state-of-the-art methods. Table I summarizes the details of all test methods. Table II lists the calibration results of the compared methods. Moreover, Fig. 2 depicts the accuracy of the compared methods after calibration. Fig. 3 shows their training curves and the number of iteration rounds. Based on the experimental results, we summarize as follows:

a) Compared with M1-M8, M9 has the best performance in terms of robot calibration accuracy. From Fig. 2 and Table II, the RMSE, Std and Max of M9 are 0.43, 0.35 and 1.05, respectively. Compared with the most accurate method M8, its RMSE, STD and Max are 0.45, 0.36 and 1.13 respectively, then the accuracy gains are 4.44%, 2.78% and 7.08% respectively. Therefore, the proposed method is helpful to improve the robot calibration accuracy.

b) As shown in Fig. 3(a) and (b), M8's converge rate is faster than that of M6. It only takes 20 iterations to converge in RMSE. From the experimental results, we see that incorporating variable step-size term into LM algorithm can improve its convergence rate.

c) As depicted in Fig. 3(c), the computational efficiency of M8 is higher than that of M6. M8 takes 50.59s to converge in RMSE, which is 62.56% less than that of M6. Hence, incorporating variable step-size term into the LM algorithm can greatly enhance its computational efficiency. Note that, to further improve the calibration accuracy of M8, we adopt an UKF algorithm to initially calibrate the robot. However, it suffers from low computational efficiency. We plan to address this issue in the future.

TABLE II
THE PERFORMANCE COMPARISON OF VARIOUS METHODS.

| Item | RMSE(mm) | Std(mm) | Max(mm) |
|---|---|---|---|
| Before | 2.09 | 2.00 | 3.36 |
| M1 | 0.66 | 0.56 | 1.71 |
| M2 | 0.70 | 0.56 | 1.76 |
| M3 | 0.64 | 0.55 | 1.51 |
| M4 | 0.59 | 0.49 | 1.28 |
| M5 | 0.85 | 0.72 | 1.91 |
| M6 | 0.50 | 0.41 | 1.16 |
| M7 | 0.67 | 0.58 | 1.59 |
| M8 | 0.45 | 0.36 | 1.13 |
| M9 | 0.43 | 0.35 | 1.05 |

## V. CONCLUSIONS

To achieve high calibration accuracy, this work proposes a novel calibration method based on an UKF algorithm and a SLM algorithm. The experimental results demonstrate that compared with the state-of-the-art calibration methods, the proposed method achieves a higher calibration accuracy.